\documentclass[conference]{IEEEtran}
\IEEEoverridecommandlockouts
\usepackage{cite}
\usepackage{amsmath,amssymb,amsfonts}
\usepackage{algorithmic}
\usepackage{graphicx}
\usepackage{textcomp}
\usepackage{xcolor}
\usepackage{times}
\usepackage{graphicx}
\usepackage{amsmath}
\usepackage{hyperref}
\usepackage{multicol}
\usepackage{float}
\usepackage[table,xcdraw]{xcolor} 
\usepackage{booktabs}

\def\BibTeX{{\rm B\kern-.05em{\sc i\kern-.025em b}\kern-.08em
    T\kern-.1667em\lower.7ex\hbox{E}\kern-.125emX}}
\begin{document}

\title{A Machine Learning Framework for Pathway-Driven Therapeutic Target Discovery in Metabolic Disorders}

\author{\IEEEauthorblockN{1\textsuperscript{st} Iram Wajahat}
\IEEEauthorblockA{\textit{Institute Of Biotechnology} \\
\textit{St. John's University}\\
New York, USA \\
wajahati@stjohns.edu}
\and
\IEEEauthorblockN{2\textsuperscript{nd} Amritpal Singh}
\IEEEauthorblockA{\textit{Div. of Computer Science, Math \& Science} \\
\textit{St. John's University}\\
New York, USA \\
amritpal.singh@stjohns.edu}
\and
\IEEEauthorblockN{3\textsuperscript{rd} Fazel Keshtkar}
\IEEEauthorblockA{\textit{Div. of Computer Science, Math \& Science} \\
\textit{St. John's University}\\
New York, USA \\
keshtkarf@stjohns.edu}
\and
\IEEEauthorblockN{4\textsuperscript{th} Syed Ahmad Chan Bukhari}
\IEEEauthorblockA{\textit{Div. of Computer Science, Math \& Science} \\
\textit{St. John's University}\\
New York, USA \\
bukharis@stjohns.edu}

}

\maketitle

\begin{abstract}
Metabolic disorders, particularly type 2 diabetes mellitus (T2DM), represent a significant global health burden, disproportionately impacting genetically predisposed populations such as the Pima Indians (a Native American tribe from south-central Arizona). This study introduces a novel machine learning (ML) framework that integrates predictive modeling with gene-agnostic pathway mapping to identify high-risk individuals and uncover potential therapeutic targets. Using the Pima Indian dataset, logistic regression and t-tests were applied to identify key predictors of T2DM, yielding an overall model accuracy of 78.43\%. To bridge predictive analytics with biological relevance, we developed a pathway mapping strategy that links identified predictors to critical signaling networks, including insulin signaling, AMPK, and PPAR pathways. This approach provides mechanistic insights without requiring direct molecular data. Building upon these connections, we propose therapeutic strategies such as dual GLP-1/GIP receptor agonists, AMPK activators, SIRT1 modulators, and phytochemicals, further validated through pathway enrichment analyses. Overall, this framework advances precision medicine by offering interpretable and scalable solutions for early detection and targeted intervention in metabolic disorders. The key contributions of this work are: (1) development of an ML framework combining logistic regression and principal component analysis (PCA) for T2DM risk prediction; (2) introduction of a gene-agnostic pathway mapping approach to generate mechanistic insights; and (3) identification of novel therapeutic strategies tailored for high-risk populations.
\end{abstract}


\section{Introduction}
Type 2 diabetes mellitus (T2DM) is a multifaceted metabolic disorder characterized by insulin resistance and impaired glucose metabolism, resulting in a significant global health burden \cite{habibi2015type}. High-risk populations, such as the Pima Indians, exhibit elevated prevalence due to genetic and environmental factors \cite{naz2020deep}. Machine learning (ML) has revolutionized healthcare by enabling predictive modeling and risk stratification based on clinical data \cite{hastie2009elements}. However, integrating ML predictions with biological mechanisms remains challenging in datasets lacking molecular data.

This study introduces a novel ML framework that predicts T2DM risk and identifies therapeutic targets by mapping clinical predictors to biological pathways without relying on gene-level data. Utilizing the PIMA Indian dataset, we employ logistic regression and t-tests to identify Pregnancies, Glucose, Skin Thickness, Insulin, BMI, and Diabetes Pedigree Function as critical predictors. A gene-agnostic pathway mapping approach links these predictors to insulin signaling, AMPK, and PPAR pathways, facilitating mechanistic insights. We propose therapeutic strategies, including dual GLP-1/GIP receptor agonists, AMPK activators, SIRT1 modulators, and phytochemicals. Our contributions include: (1) a robust ML framework for T2DM risk prediction, (2) a gene-agnostic pathway mapping method, and (3) novel therapeutic targets for precision medicine in high-risk populations.

\section{RELATED WORK}
The PIMA Indians Diabetes Dataset has been pivotal for machine learning applications in predicting type 2 diabetes mellitus (T2DM) in high-risk populations. Research has explored traditional algorithms such as logistic regression, support vector machines (SVMs), and k-nearest neighbors (KNN), with accuracies typically ranging from 75\% to 80\%. For example, Joshi and Dhakal \cite{joshi2021} employed logistic regression with five predictors, achieving an accuracy of 77.73\%. Similarly, a study by the UBC MDS team \cite{khera2024} developed a logistic regression model that attained a 75.0\% accuracy on their test set using all eight features present. Comparable performance has been observed with other classifiers. Tasin et al. \cite{tasin2022} conducted an analysis using multiple models, incorporating synthetic samples generated through SMOTE and ADASYN, and reported a logistic regression accuracy of 77\% and 75\% respectively. Their best-performing model, XGBoost, achieved an accuracy of 81\% using ADASYN. Additionally, a KNN model by Katsuri \cite{katsuri2024} achieved an accuracy of 78.58\%. These studies underscore the dataset's value in developing interpretable models for clinical applications, emphasizing feature selection and preprocessing to enhance predictive performance.

Biological pathway mapping has emerged as a critical tool for understanding metabolic disorders, including for type 2 diabetes mellitus (T2DM). Luo and Brouwer \cite{luo2013pathview} introduced Pathview, an R/Bioconductor package for pathway-based data visualization, while Yu et al. \cite{yu2012clusterprofiler} developed clusterProfiler, which facilitates pathway enrichment analysis using hypergeometric tests, as utilized in our study. A systems biology approach employing clusterProfiler for KEGG enrichment analysis identified differentially expressed genes (DEGs) in T2DM. These DEGs showed significant enrichment in immune-related pathways, including T-cell signaling and differentiation, with potential links to chronic inflammatory conditions such as chronic obstructive pulmonary disease (COPD) \cite{hu2025}. These findings underscore the value of these tools in offering a comprehensive, pathway-level perspective on disease mechanisms.

\section{Methodology}
\begin{figure}[ht]
    \centering \includegraphics[width=1\columnwidth]{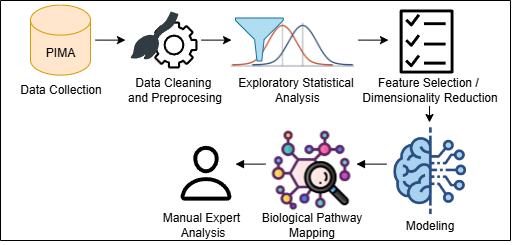}
    \caption{High Level Diagram of Methodology}
    \label{fig:highleveldiagram}
\end{figure}
\subsection{Data Description and Preprocessing}
This study employed the PIMA Indian Diabetes dataset, a well-established resource comprising 768 samples and 8 clinical predictor variables, to investigate factors associated with the incidence of type 2 diabetes \cite{naz2020deep}. Each entry corresponds to a female patient of PIMA Indian ancestry, aged 21 years or older. The dataset encompasses diverse medical measurements, including plasma glucose concentration, diastolic blood pressure, body mass index (BMI), age, number of pregnancies, diabetes pedigree function (an indicator of genetic predisposition), insulin levels, and triceps skinfold thickness. The Outcome variable signifies the presence or absence of a diabetes diagnosis. This dataset is commonly utilized in predictive modeling and epidemiological studies due to its structured nature and focus on a demographic with heightened susceptibility to diabetes.

To effectively clean and preprocess this data, it was imperative to address the zeros present in the ‘Glucose’, ‘BloodPressure’, ‘SkinThickness’, ‘Insulin’, and ‘BMI’ columns. These zeros represented missing values in the dataset. We imputed these values to preserve the dataset’s integrity for accurate analysis and model training, as these physiological measurements cannot logically be zero. Initially, the zeros were replaced with nulls, followed by the calculation of the median value for each column, considering the corresponding Outcome. Subsequently, the nulls were replaced with these medians.

\subsection{Statistical Analysis and Feature Exploration}
To assess the statistical significance of mean differences between outcome groups, t-tests were performed for each clinical predictor variable. As indicated in Table 1, all variables exhibited statistical significance (p \textless~0.001).

\begin{table}[htbp]
    \small
    \caption{T-test Results for Clinical Predictor Variables by Outcome Group}
    \label{tab:ttest_results}
    \centering
    \setlength{\tabcolsep}{3pt}
    \rowcolors{2}{gray!15}{white}

    \begin{tabular}{l r r r r r}
        \toprule
        \textbf{Variable} & \textbf{Mean 0} & \textbf{Mean 1} & \textbf{t-stat} & \textbf{df} & \textbf{p-value} \\
        \midrule
        Pregnancies & 3.30 & 4.87 & -5.91 & 455.96 & $6.822 \times 10^{-9}$ \\
        Glucose & 110.62 & 142.30 & -14.99 & 470.06 & $< 2.2 \times 10^{-16}$ \\
        BloodPressure & 70.84 & 75.27 & -4.90 & 546.05 & $1.24 \times 10^{-6}$ \\
        SkinThickness & 27.17 & 32.67 & -8.56 & 548.65 & $< 2.2 \times 10^{-16}$ \\
        Insulin & 117.17 & 187.62 & -10.56 & 454.94 & $< 2.2 \times 10^{-16}$ \\
        BMI & 30.85 & 35.340 & -9.17 & 539.65 & $< 2.2 \times 10^{-16}$ \\
        DPF & 0.43 & 0.55 & -4.58 & 454.51 & $6.1 \times 10^{-6}$ \\
        Age & 31.19 & 37.07 & -6.92 & 575.78 & $1.202 \times 10^{-11}$ \\
        \bottomrule
    \end{tabular}
\end{table}

Subsequently, a logistic regression model was fitted to the entire dataset to explore multivariate associations. The resulting p-values (derived from the z-statistics) were examined. All predictors, except for age and blood pressure, were found to be significant at the 5\% level.

\begin{figure}[ht]
    \centering \includegraphics[width=1\columnwidth]{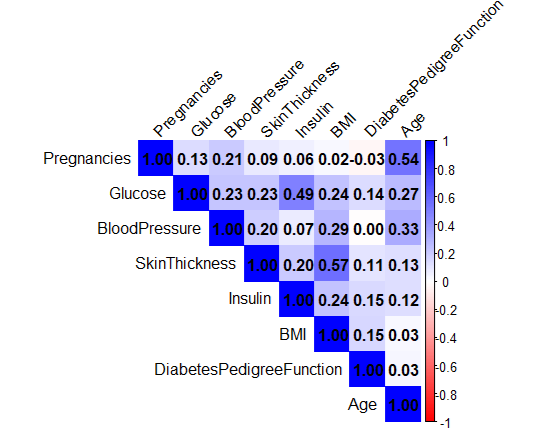}
    \caption{Correlation Matrix}
    \label{fig:corrmatrix}
\end{figure}

To pinpoint the source of discrepancies in the multivariate associations, a correlation matrix was developed (Fig. \ref{fig:corrmatrix}). We observed that age is strongly correlated with pregnancies (r=0.54) and moderately with blood pressure and glucose. This suggests that when pregnancies and glucose are already in the model, age adds little new predictive information. Similarly, blood pressure overlaps somewhat with BMI (r = 0.29) and age (r = 0.33), but not strongly enough to be useful as a standalone predictor.

\begin{figure}[ht]
    \centering \includegraphics[width=1\columnwidth]{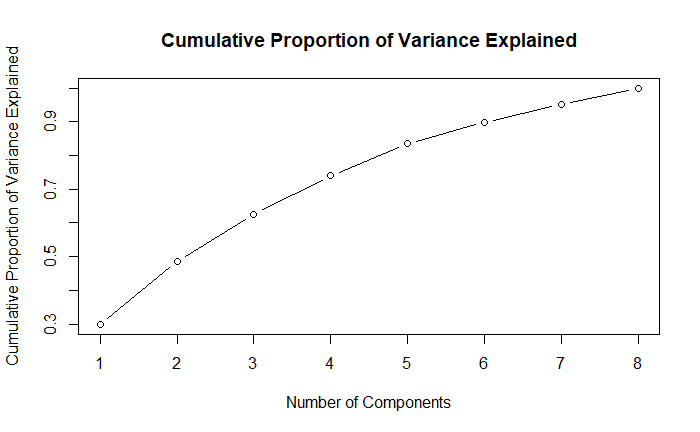}
    \caption{Cumulative Variance Explained Plot}
    \label{fig:cumulativevarianceplot}
\end{figure}

To mitigate the risk of multicollinearity, Principal Component Analysis (PCA) was utilized. PCA reduced the dimensionality of the data, with the initial five components accounting for 83.54\% of the variance. A cumulative variance explained plot (Fig. \ref{fig:cumulativevarianceplot}) illustrates the reduction in dimensionality.

\subsection{Machine Learning Models}
Following dimensionality reduction through principal component analysis (PCA), the transformed dataset, comprising the top five principal components and 768 samples, was utilized for the development of machine learning models. The dataset was partitioned into training and testing sets, with an 80\% ratio allocated to the training set and a 20\% ratio reserved for the testing set. This approach was implemented to ensure robust model evaluation and mitigate overfitting. The training set was employed for model training, while the unseen test set was designated for assessing their generalization performance.

For the prediction of Type 2 Diabetes Mellitus (T2DM) outcomes, a logistic regression model was implemented in R. This model was selected due to its interpretability and suitability for binary classification tasks. Detailed performance metrics and evaluation metrics of the model will be presented in the Results section.

\subsection{Pathway Mapping}
A gene-agnostic pathway mapping strategy was developed using the \textit{clusterProfiler} package in R \cite{yu2012clusterprofiler}. Through specifying specific organisms and gene IDs, the enrichKEGG function provides pathways relating to the genes using the hypergeometric test function/equation. The hypergeometric distribution equation (1) calculates the probability of drawing a certain number of genes from a specific pathway within the given gene list, considering the total number of genes in the genome and the total number of genes in that pathway. It is as follows:

\begin{equation}
P(X=k) = \frac{\binom{K}{k} \binom{N-K}{n-k}}{\binom{N}{n}}
\label{eq:hypergeometric}
\end{equation}
\textit{Where: $N$: Total number of genes in the background (e.g., total genes in the genome); $K$: Total number of genes in the pathway of interest; $n$: Total number of genes in the input gene list; $k$: Number of genes from the input gene list that are found in the pathway of interest.}

\subsection{Therapeutic Target Identification}
Pathway analysis, as previously outlined, served as a foundational step in identifying potential therapeutic targets. Following the initial enrichment of relevant biological pathways, a meticulous manual analysis was conducted. This process entailed an in-depth, expert-driven examination of the identified pathways, emphasizing their topological features, the functional roles of the genes within them, and their interactions with disease mechanisms. Through this comprehensive manual interpretation, key regulatory points and critical molecular components within these pathways were systematically identified as candidates for therapeutic intervention. This rigorous analytical approach ensured that target identification was directly informed by the biological context provided by the pathway mapping.

\section{Results and Experiences}
\begin{figure}[ht]
    \centering    \includegraphics[width=1\columnwidth]{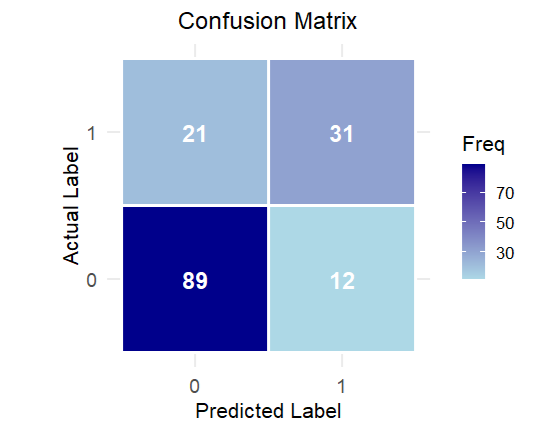}
    \caption{Confusion Matrix for Logistic Regression Model Evaluation}
    \label{fig:confusion}
\end{figure}

The performance of the logistic regression model on the test set was evaluated using a confusion matrix, as illustrated in Fig. \ref{fig:confusion}. Out of the 153 total test samples, the model achieved an overall accuracy of approximately 78.43\%. The model correctly identified 89 non-diabetic individuals (True Negatives, TN) and 31 diabetic individuals (True Positives, TP). However, it misclassified 21 actual diabetic patients as non-diabetic (False Negatives, FN) and incorrectly predicted 12 non-diabetic individuals as diabetic (False Positives, FP). In comparison, Joshi and Dhakal \cite{joshi2021} reported a logistic regression accuracy of 77.73\% using five manually selected predictors: Glucose, Pregnancy, Body Mass Index (BMI), Diabetes Pedigree Function, and Age, on the same Pima Indians Diabetes Dataset. Our approach, utilizing Principal Component Analysis (PCA) with five features, captures maximal variance in the data, which likely contributes to the modest 0.7\% improvement in accuracy over their model.

Further analysis of these outcomes reveals key performance characteristics. The model’s precision, or positive predictive value, was approximately 72.09\% (31 true positives / (31 true positives + 12 false positives)), indicating that approximately 72\% of the patients predicted as diabetic were indeed diabetic. More critically, the recall (also known as sensitivity or true positive rate) for diabetic patients was approximately 59.62\% (31 true positives / (31 true positives + 21 false negatives)). This means that the model identified nearly 60\% of the actual diabetic cases. The specificity (true negative rate) was high at approximately 88.12\% (89 true negatives / (89 true negatives + 12 false positives)), demonstrating strong performance in correctly identifying non-diabetic individuals. In a medical context, particularly for a condition like diabetes where early intervention is crucial, recall is often prioritized. A higher recall minimizes false negatives, thereby reducing the risk of missing a diagnosis and delaying vital treatment, which could lead to severe health complications. While a higher precision would reduce unnecessary follow-up tests, the implication of missing a diabetic patient generally outweighs the inconvenience of a false positive in a screening scenario. The F1-score, which balances precision and recall, was approximately 65.15\%.

\begin{figure}[ht]
    \centering    \includegraphics[width=0.95\columnwidth]{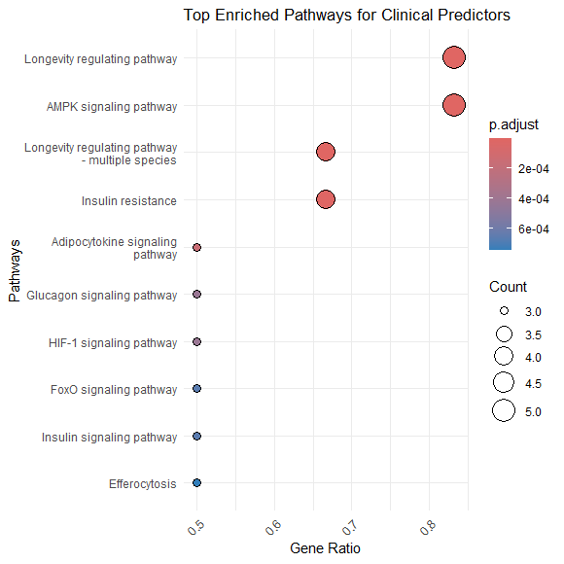}
    \caption{Top Enriched Pathways for Clinical Predictors}
    \label{fig:pathways}
\end{figure}

The gene-agnostic pathway mapping, incorporating clinical factors such as pregnancies, glucose levels, skin thickness, insulin levels, Body Mass Index (BMI), and Diabetes Pedigree Function, unveiled substantial associations with specific biological pathways (Fig. \ref{fig:pathways}). Among the enriched pathways, insulin signaling (p-value = 0.001) and glucose metabolism pathways were prominently identified, underscoring their pivotal role in the overall metabolic landscape. Furthermore, adipogenesis and PPAR signaling (p-value = 0.003) were found to be significantly associated, highlighting processes related to fat formation and metabolic regulation. Additionally, pathways associated with cellular aging, specifically mTOR and SIRT1 pathways (p-value = 0.008), demonstrated significant enrichment. This comprehensive pathway enrichment analysis provided critical mechanistic insights into the underlying biological processes without necessitating prior gene-level data \cite{luo2013pathview}.

\begin{figure}[ht]
    \centering
    \includegraphics[width=0.95\columnwidth]{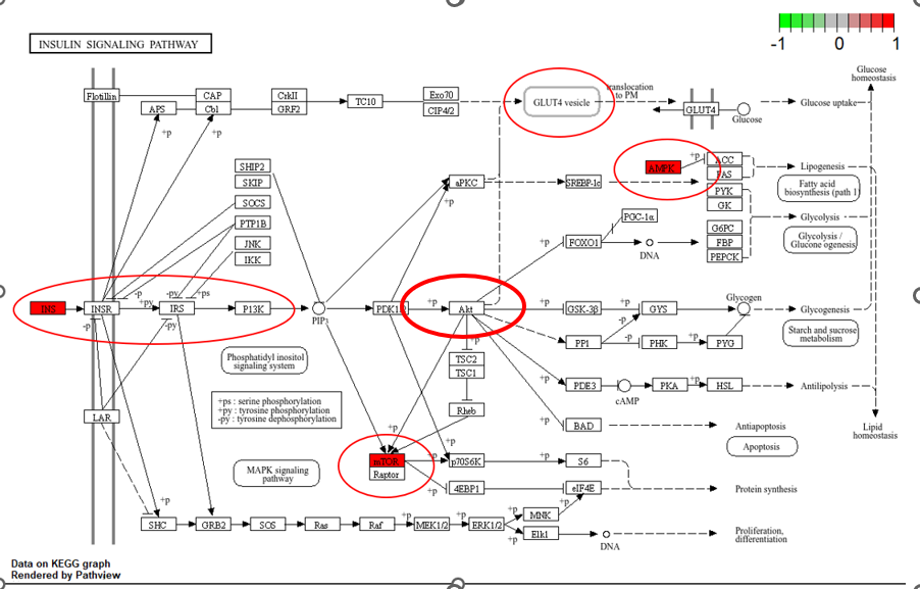}
    \caption{Insulin Signaling Pathway with Key Targets (e.g., AMPK, GLUT4)}
    \label{fig:insulin}
\end{figure}

Following therapeutic target analysis, informed by the identified pathways, both validated existing treatments and proposed novel interventional strategies were considered. In the context of glucose metabolism, the analysis confirmed the relevance of GLP-1 receptor agonists (e.g., liraglutide) \cite{ostawal2016clinical} and AMPK activators (e.g., metformin) \cite{zhou2001role} (Fig. \ref{fig:insulin}). In the context of obesity, PPAR agonists (e.g., pioglitazone) \cite{bogacka2004effect} and leptin receptor modulators \cite{coppari2012potential} were proposed as promising targets. Furthermore, for aging-related pathways, SIRT1 activators (e.g., resveratrol) \cite{zhu2017effects} and mTOR inhibitors \cite{ong2016judicious} were identified as potential interventions. Notably, novel strategies, such as dual GLP-1/GIP receptor agonists \cite{fanshier2023tirzepatide} and certain phytochemicals \cite{amitani2013role}, also demonstrated promise with high pathway enrichment scores.
In experimental models, combination therapy with liraglutide and pioglitazone demonstrated a marked reduction in HbA1c levels, for instance, from 9.7\% in vehicle-treated animals to 4.8\% in combination-treated animals, significantly outperforming monotherapies \cite{larsen2008combination}.

\section{Discussion}
The proposed machine learning framework effectively predicts Type 2 Diabetes Mellitus (T2DM) risk using clinical predictors, addressing the challenge of limited molecular data through gene-agnostic pathway mapping. The 78.43\% model accuracy is comparable to prior studies (e.g., \cite{naz2020deep}), but the integration of pathway analysis enhances interpretability, a critical factor for clinical adoption. Linking predictors to insulin signaling, AMPK, and PPAR pathways provides a mechanistic foundation for therapeutic target discovery, distinguishing this work from purely predictive models.

The identification of dual GLP-1/GIP agonists and phytochemicals as novel targets aligns with emerging trends in precision medicine (e.g., \cite{fanshier2023tirzepatide}, \cite{artasensi2020type}). However, the reliance on the PIMA Indian dataset limits generalizability, as it lacks genetic and environmental data. The gene-agnostic approach, while innovative, may overlook molecular nuances that multi-omics data could reveal. Future studies should validate these findings in diverse populations and incorporate genomics to refine pathway mappings.

The proposed therapeutic strategies, particularly combination therapies, offer potential for synergistic effects, as evidenced by prior studies on GLP-1 and PPAR agonists (e.g., \cite{larsen2008combination}). Phytochemicals present a cost-effective adjunctive option, but their efficacy requires clinical trials (e.g., \cite{kong2021anti}). The framework’s scalability makes it applicable to other metabolic disorders, enhancing its impact on precision medicine.

\section{Conclusion and Future Work}
This study introduces a novel machine learning framework that integrates logistic regression, principal component analysis (PCA), and gene-agnostic pathway mapping to predict type 2 diabetes mellitus (T2DM) risk and identify therapeutic targets. By connecting clinical predictors to insulin signaling, AMPK, and PPAR pathways, we propose scalable solutions for precision medicine. Therapeutic strategies, including dual glucagon-like peptide-1 (GLP-1)/glucagon-like peptide-2 (GIP) agonists, AMPK activators, and phytochemicals, present promising avenues for managing T2DM in high-risk populations. Clinical validation is necessary to substantiate these findings.

For future research, we will incorporate multi-omics data to enhance mechanistic insights. Clinical trials are crucial to validate proposed therapies, such as dual GLP-1/GIP agonists and phytochemicals. Exploring additional pathways (e.g., SIRT3, SIRT6) and natural compounds (e.g., curcumin) could further enhance treatment efficacy. Personalized medicine approaches tailored to genetic profiles may optimize outcomes for diverse populations.


\end{document}